\documentclass[twocolumn]{article}

\usepackage[width=17cm,height=22cm]{geometry}
\usepackage[english]{babel}
\usepackage[utf8]{inputenc}
\usepackage{fancyvrb}
\usepackage{authblk}
\usepackage{hyperref}
\usepackage{tcolorbox}
\usepackage{slashbox}

\usepackage{float}

\usepackage{tabularx}
\newcolumntype{C}{>{\centering}X}

\usepackage{tikz}
\usetikzlibrary{shapes.geometric}

\title{ReviewQA: a relational aspect-based opinion reading dataset}
\author[1]{Quentin Grail}
\author[1]{Julien Perez}
\affil[1]{NAVER LABS Europe}
\affil[ ]{\texttt{\{quentin.grail, julien.perez\}@naverlabs.com}}

\date{}

\usepackage{xspace}
\newcommand{\system}[1]{\texttt{#1}\xspace}
\newcommand{\memnn}{\system{MemN2N}}

\newcommand\score[2]{
\pgfmathsetmacro\pgfxa{#1+1}
\tikzstyle{scorestars}=[star, star points=5, star point ratio=2.25, draw,inner sep=1.3pt,anchor=outer point 3]
  \begin{tikzpicture}[baseline]
    \foreach \i in {1,...,#2} {
    \pgfmathparse{(\i<=#1?"yellow":"gray")}
    \edef\starcolor{\pgfmathresult}
    \draw (\i*1.75ex,0) node[name=star\i,scorestars,fill=\starcolor]  {};
   }
  \end{tikzpicture}
}

\begin{document}

\maketitle

\begin{abstract}
Deep reading models for question-answering have demonstrated promising performance over the last couple of years. However current systems tend to learn how to cleverly extract a span of the source document, based on its similarity with the question, instead of seeking for the appropriate answer. Indeed, a reading machine should be able to detect relevant passages in a document regarding a question, but more importantly, it should be able to reason over the important pieces of the document in order to produce an answer when it is required. To motivate this purpose, we present ReviewQA, a question-answering dataset based on hotel reviews. The questions of this dataset are linked to a set of relational understanding competencies that we expect a model to master. Indeed, each question comes with an associated type that characterizes the required competency. With this framework, it is possible to benchmark the main families of models and to get an overview of what are the strengths and the weaknesses of a given model on the set of tasks evaluated in this dataset. Our corpus contains more than 500.000 questions in natural language over 100.000 hotel reviews. Our setup is projective, the answer of a question does not need to be extracted from a document, like in most of the recent datasets, but selected among a set of candidates that contains all the possible answers to the questions of the dataset. Finally, we present several baselines over this dataset.
\end{abstract}
 

\section{Introduction}
\label{sec:Introduction}

A large majority of the human knowledge is recorded through text documents.
That is why ability for a system to automatically infer information from text without any structured data has become a major challenge.
Answering questions about a given document is a relevant proxy task that has been proposed as a way to evaluate the reading ability of a given model.
In this configuration, a text document such as a news article, a document from Wikipedia or any type of text is presented to a machine with an associated set of questions. The system is then expected to answer these questions and evaluated by its accuracy on this task. The machine reading framework is very general and we can imagine a large panel of questions that can possibly handle most of the standard natural language processing tasks. For example, the task of named entities recognition can be formulated as a machine reading one where your document is the sentence and the question would be 'What are the named entities mentioned in this sentence?'. These natural language interactions are an important objective for reading systems.
\\

Recently, many datasets have been proposed to build and evaluate reading models \cite{Rajpurkar2016SQuAD10, Trischler2017NewsQAAM}. From cloze style questions \cite{DBLP:journals/corr/abs-1711-05073} to open questions \cite{Chen2017ReadingWT}, from synthetic data \cite{Weston2015TowardsAQ} to human written articles  \cite{Hermann2015TeachingMT}, many styles of documents and questions have been proposed to challenge reading models. 
The correct answer to the questions proposed in most of these datasets is a span of text of the source document, which can be restricted to a single word in several cases.
It means that the answer should explicitly be present in the source document and that the model should be able to locate it.\\

Different models have already shown \textit{superhuman} performance on several of these datasets and particularly on the SQuAD dataset composed of Wikipedia articles \cite{DBLP:journals/corr/HuPQ17, wei2018fast}.
However, some limits of such models have been highlighted when they encounter perturbations into the input documents \cite{Jia2017AdversarialEF}. 
Indeed almost all of the state of the art models on the SQuAD dataset suffer from a lack of robustness against adversarial examples.
Once the model is trained, a meaningless sentence added at the end of the text document can completely disturb the reading system. 
Conversely, these adversarial examples do not seem to fool a human reader who will be capable of answering the questions as well as without this perturbation. 
One possible explanation of this phenomenon is that computers are good at extracting patterns in the document that match the representation of the question. If multiple spans of the documents look similar to the questions, the reader might not be able to decide which one is relevant.
Moreover, Wikipedia articles tend to be written with the same standard writing style, factual, unambiguous. Such writing style tends to favor the pattern matching between the questions and the documents. 
This format of documents/questions has certainly influenced the design of the comprehension models that have been proposed so far. Most of them are composed of stacked attention layers that match question and document representations.\\

Following concepts proposed in the 20 bAbI tasks \cite{Weston2015TowardsAQ} or in the visual question-answering dataset CLEVR \cite{Johnson2017CLEVRAD}, we think that the challenge, limited to the detection of relevant passages in a document, is only the first step in building systems that truly understand text. The second step is the ability of reasoning with the relevant information extracted from a document.
To set up this challenge, we propose to leverage on a hotel reviews corpus that requires reasoning skills to answer natural language questions.
The reviews we used have been extracted from TripAdvisor and originally proposed in \cite{Wang2010LatentAR, Wang2011LatentAR}. In the original data, each review comes with a set of rated aspects among the seventh available: \textit{Business service, Check in / Front Desk, Cleanliness, Location, Room, Sleep Quality, Value} and for all the reviews an \textit{Overall} rating. In this articles we propose to exploit these data to create a dataset of question-answering that will challenge 8 competencies of the reader.\\

Our contributions can be summarized as follow:

\begin{itemize}
\item We propose to evaluate the sentiment analysis task directly through the more general framework of question-answering.

\item Based on hotel reviews from TripAdvisor, we propose a set of 8 reasoning tasks that a reading model should master.

\item We release ReviewQA, a large question-answering dataset that evaluates these 8 tasks through crowdsourced and backtranslated natural language questions.

\item Finally, we propose 4 baselines on this dataset including a novel model inspired by one of the state-of-the-art extractive reading models \cite{Wang2017GatedSN}.

\end{itemize}

\section{Related work}
\label{sec:Related work}

\subsection{Machine comprehension datasets}

ReviewQA is proposed as a novel dataset regarding the collection of the existing ones. Indeed a large panel of available datasets, that evaluate models on different types of documents, can only be valuable for designing efficient models and learning protocols. In this following part, we describe several of these datasets.\\

\noindent
\textbf{SQuAD: }The Standford Question Answering Dataset (SQuAD) introduced in \cite{Rajpurkar2016SQuAD10} is a large dataset of natural questions over the 500 most popular articles of Wikipedia.
All the questions have been crowdsourced and answers are spans of text extracted from source documents.
This dataset has been very popular these last two years and the performance of the architectures that have been proposed have rapidly increased until several models surpass the human score.
Indeed, in the original paper human performance has been measured at 82.304 points for the exact match metric and at the time we are writing this paper four models have already a higher score. In another hand \cite{Jia2017AdversarialEF} has shown that these models suffer from a lack of robustness against adversarial examples that are meaningless from a human point of view. This suggests the need for a more challenging dataset that will allow developing strongest reasoning architectures.\\

\noindent
\textbf{NewsQA: }NewsQA \cite{Trischler2017NewsQAAM} is a dataset very similar to SQuAD.
It contains 120.000 human generated questions over 12.000 articles form CNN originally introduced in \cite{Hermann2015TeachingMT}. It has been designed to be more challenging than SQuAD with questions that might require to extract multiple spans of text or not be answerable.\\

\noindent
\textbf{WikiHop and MedHop: }These are two recent datasets introduced in \cite{Welbl2017ConstructingDF}. 
Unlike SQuAD and NewsQA, important facts are spread out across multiple documents and, in order to answer a question, it is necessary to jump over a set of passages to collect the required information. 
The relevant passages are not explicitly mentioned in the data so this dataset measures the ability that a model has to navigate across multiple documents.
The questions come with a set of candidates which are all present in the text.\\

\noindent
\textbf{MS Marco: }This dataset has been released in \cite{Nguyen2016MSMA}.
The documents come from the internet and the questions are real user queries asked through the bing search engine.
The dataset contains around 100.000 queries and each of them comes with a set of approximatively 10 relevant passages. Like in SQuAD, several models are already doing \textit{superhuman} performances on this dataset.\\

\noindent
\textbf{Facebook bAbI tasks: }This is a set of 20 toy tasks proposed in \cite{Weston2015TowardsAQ} and designed to measure text understanding. Each task requires a certain capability to be completed like induction, deduction and more. Documents are synthetic stories, composed of few sentences that describe a set of actions. This dataset was one of the first attempt to introduce a general set of prerequisite capabilities required for the reading task. Although it has been a very challenging framework, beneficial to the emergence of the attention mechanism inside the reading architectures, a Gated end-to-end memory network \cite{Liu2017GatedEM} now succeed in almost all of the 20 tasks. One of the possible reason is that the data are synthetic data, without noise or ambiguity. We propose a comparable framework with understanding and reasoning tasks based on user-generated comments that are much more realistic and that required language competencies to be understood. 
\\

\textbf{CLEVR: } Beyond textual question-answering, Visual Question-Answering (VQA) has been largely studied during the last couple of years. More recently, the problem of relational reasoning has been introduced through this dataset \cite{Johnson2017CLEVRAD}. The main original idea was to introduce relational reasoning questions over object shapes and placements. This dataset has already motivated the development of original deep models. To the best of our knowledge, no natural language question-answering corpus has been designed to investigate such capabilities. As we will present in the following of this paper, we think sentiment analysis is particularly suited for this task and we will introduce a novel machine reading corpus with such capability requirements.

\subsection{Attention-based models for aspect-based sentiment analysis}

Sentiment analysis is one of the historical tasks of Natural Language Processing. 
It is an important challenge for companies, restaurants, hotels that aim to analyze customer satisfaction regarding products and quality of services.
Given a text document, the objective is to predict its overall polarity.
Generally, it can be positive, negative or neutral.
This analysis gives a quick overview of a general sentiment over a set of documents, but this framework tends to be restrictive. Indeed, one document tends to express multiple opinions of different aspects. For instance, in the sentence: \textit{The fish was very good but the service was terrible}, there is not a general dominant sentiment, and a finer analysis is needed.
The task of aspect-based sentiment analysis aims to predict a polarity of a sentence regarding a given aspect. In the previous example a positive polarity should be associated to the aspect \textit{food}, and on the contrary, a negative sentiment is expressed regarding the \textit{quality of the service}.\\

The idea of using models originally designed for question-answering, for the sentiment analysis task has been introduced in \cite{Tay2017DyadicMN, Tang2016AspectLS}. In these papers, several adaptations of the end-to-end memory network (\memnn) \cite{Sukhbaatar2015EndToEndMN} are used to predict the polarity of a review regarding a given aspect. In that configuration, the review is encoded into the memory cells and the controller, usually initialized with a representation of the question, is initialized with a representation of the aspect. The analysis of the attention between the values of the controller and the document has shown interesting results, by highlighting relevant part of a document regarding an aspect.

\section{ReviewQA dataset}
\label{sec:Dataset}

We think that evaluating the task of sentiment analysis through the setup of question-answering is a relevant playground for machine reading research. Indeed natural language questions about the different aspects of the targeted venues are typical kind of questions we want to be able to ask to a system.
In this context, we introduce a set of reasoning questions types over the relationships between aspects.
We propose ReviewQA, a dataset of natural language questions over hotel reviews. These questions are divided into 8 groups, regarding the competency required to be answered. In this section, we describe each task and the process followed to generate this dataset.

\subsection{Original data}

\begin{figure}
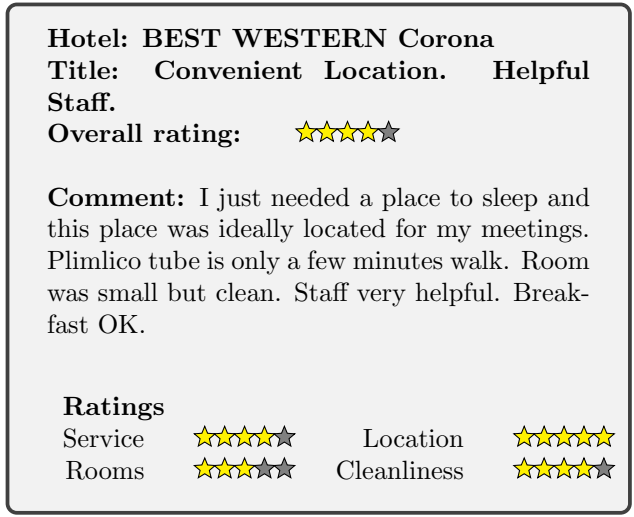

\begin{tcolorbox}
\textbf{Hotel: BEST WESTERN Corona}\\
\textbf{Title: Convenient Location. Helpful Staff.}\\
\textbf{Overall rating: \score{4}{5}}\\

\textbf{Comment: }I just needed a place to sleep and this place was ideally located for my meetings. Plimlico tube is only a few minutes walk. Room was small but clean. Staff very helpful. Breakfast OK.

\begin{center}
\begin{tabular}{rlrlrl}
\multicolumn{4}{l}{\textbf{Ratings}}\\
Service & \score{4}{5} & Location & \score{5}{5}\\
Rooms & \score{3}{5} & Cleanliness & \score{4}{5}\\
\end{tabular}
\end{center}
\end{tcolorbox}
\caption{An example from the original dataset.}
\label{data_example}
\end{figure}

We used a set of reviews extracted from the TripAdvisor website and originally proposed in \cite{Wang2010LatentAR} and \cite{Wang2011LatentAR}. This corpus is available at \url{http://www.cs.virginia.edu/~hw5x/Data/LARA/TripAdvisor/TripAdvisorJson.tar.bz2}.
Each review comes with the name of the associated hotel, a title, an overall rating, a comment and a list of rated aspects. 
From 0 to 7 aspects, among \textit{value, room, location, cleanliness, check-in/front desk, service, business service}, can possibly be rated in a review. Figure \ref{data_example} displays a review extracted from this dataset.

\subsection{Relational reasoning competencies}

\textbf{Objective:} 
Starting with the original corpus, we aim at building a machine reading task where natural language questions will challenge the model on its understanding of the reviews. 
Indeed learning relational reasoning competencies over natural language documents is a major challenge of the current reading models. These original raw data allow us to generate relational questions that can possibly require a global understanding of the comment and reasoning skills to be treated. For example, asking a question like \textit{What is the best aspect rated in this comment ?} is not an easy question that can be answered without a deep understanding of the review. It is necessary to capture all the aspects mentioned in the text, to predict their rating and finally to select the best one.
The tasks and the dataset we propose are publicly available at \url{http://www.europe.naverlabs.com/Blog/ReviewQA-A-novel-relational-aspect-based-opinion-dataset-for-machine-reading}  \\

We introduce a list of 8 different competencies that a reading system should master in order to process reviews and text documents in general. 
These 8 tasks require different competencies and a different level of understanding of the document to be well answered. 
For instance, detecting if an aspect is mentioned in a review will require less understanding of the review than predicting explicitly the rating of this aspect. 
Table \ref{tasks} presents the 8 tasks we have introduced in this dataset with an example of a question that corresponds to each task. We also provide the expected \textit{type} of the answer (Yes/No question, rating question...). It can be an additional tool to analyze the errors of the readers.\\

\begin{table*}
\small
\begin{tabular}{c|p{7cm}p{4cm}p{2cm}}

Task id & Description/Comment & Example & Expected answer\\
\hline
\hline\\

1 & \textbf{Detection of an aspect in a review.} This is the very fundamental task. 
Its objective is to measure how well a model is able to detect whether an aspect is mentioned or not in a review. & Is sleep quality mentioned in this review ? & Yes/No\\

\hline\\

2 & \textbf{Prediction of the customer general satisfaction.} This second task measures how well a model is able to predict the overall positivity or negativity of a given review. & Is the client satisfy by this hotel ? & Yes/No\\

\hline\\

3 & \textbf{Prediction of the global trend of an aspect in a given review.} This task measures the satisfaction of a client per aspect. This is a precision over the last task since a client can be globally satisfied by a hotel but not satisfied regarding a certain aspect. & Is the client satisfied with the cleanliness of the hotel ? & Yes/No\\

\hline\\

4 & \textbf{Prediction of whether the rating of a given aspect is above or under a given value.} This evaluate more precisely how the reader is able to infer the ration of an aspect & Is the rating of location under 4 ? & Yes/No\\

\hline\\

5 & \textbf{Prediction of the exact rating of an aspect in a review.} This task measures precisely the satisfaction of a client regarding an aspect. This is the finest measure which can be extracted from the review. & What is the rating of the aspect Value in this review ? &  A rating between 1 and 5\\

\hline\\

6 & \textbf{Prediction of the list of all the positive/negative aspects mentioned in the review.} \newline To answer a question of this type, the system needs to detect all the aspects that are mentioned in the review and their associated polarity. This question measures the capability of a model to filter positive and negative information.&Can you give me a list of all the positive aspects in this review ?  & a list of aspects\\

\hline\\

7.0 & \textbf{Comparison between aspects.} Depending on the case, this question can require the model to understand precisely the level of satisfaction of the user regarding the two mentioned aspects. & Is the sleep quality better than the service in this hotel ? & Yes/No\\
7.1 & & Which one of these two aspects, service, location has the best rating ? & an aspect\\

\hline\\

8 & \textbf{Prediction of the strengths and weaknesses in a review.} This is probably the hardest task of the dataset. 
It requires a complete and precise understanding of the review.
To perform well on this task, a model should probably master all the previous tasks. & What is the best aspect rated in this comment ? & an aspect\\

\hline
\end{tabular}
\caption{Descriptions and examples of the 8 tasks evaluated in ReviewQA.}
\label{tasks}
\end{table*}

\subsection{Construction of the dataset}
\begin{center}
\begin{figure}
\includegraphics[scale=.5]{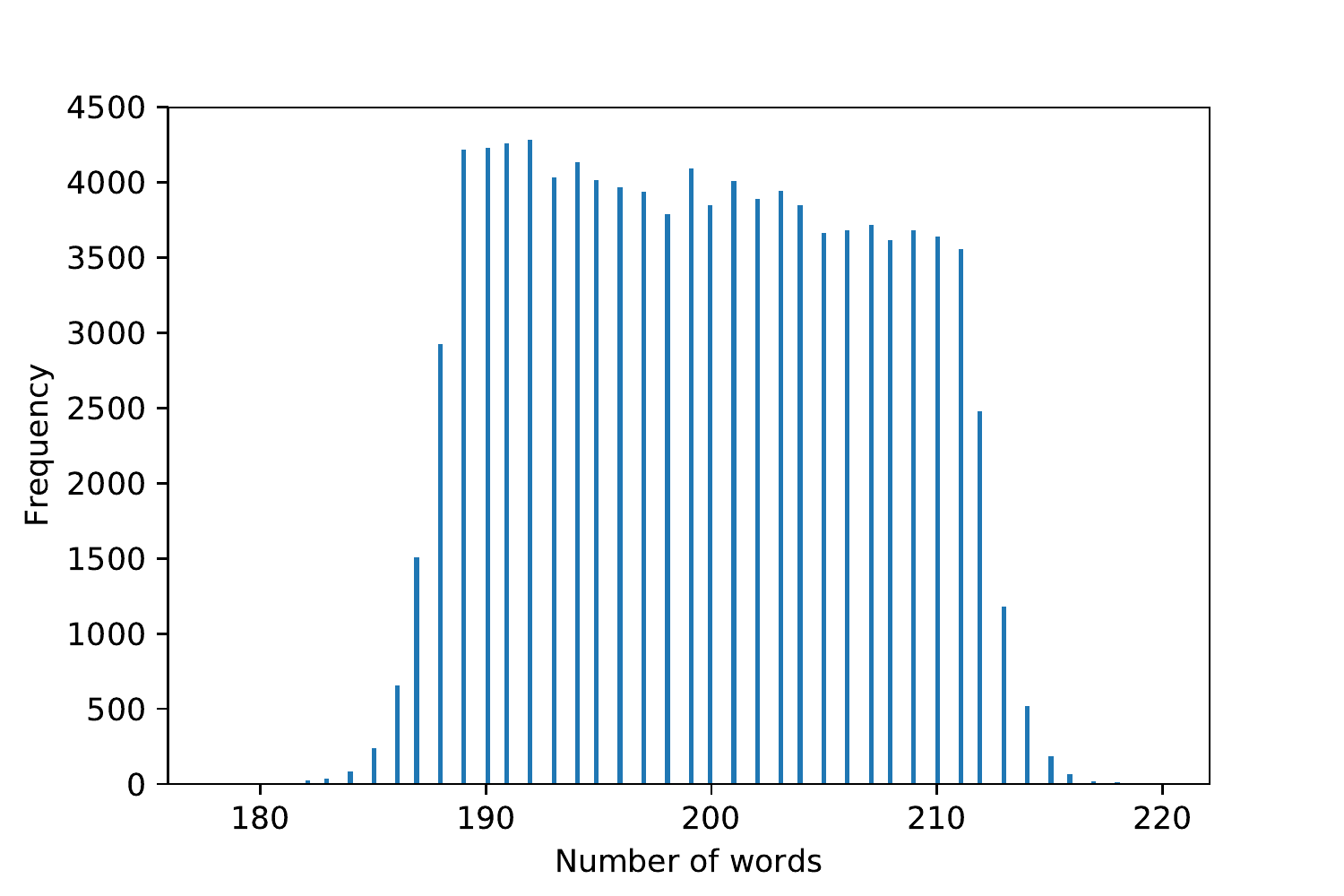}
\caption{Number of words per review.}
\label{word_distrib}
\end{figure}
\end{center}

\begin{figure}
\includegraphics[scale=.5]{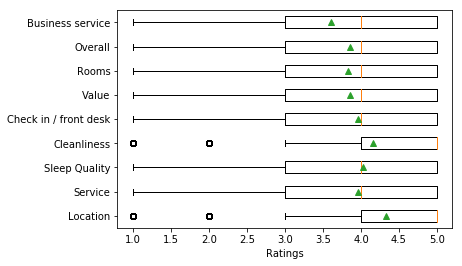}
\caption{Distribution of the ratings per aspect.}
\label{aspects}
\end{figure}

We sample 100.000 reviews from the original corpus. 
Figure \ref{word_distrib} presents the distribution of the number of words of the reviews in the dataset.
We explicitly favor reviews which contain an important number of words. In average, a review contains 200 words.
Indeed these long reviews are most likely to contain challenging relations between different aspects.
A short review which deals with only a few aspects is more likely to not be very relevant to the challenge we want to propose in this dataset. 
Figure \ref{aspects} displays the distribution of the ratings per aspects in the 100.000 reviews we based our dataset. We can see that the average values of these ratings tend to be quite high. It could have introduced bias if it was not the case for all the aspects. For example, we do not want that the model learns that in general, the service is rated better than the location and them answer without looking at the document. Since this situation is the same for all the aspects, the relational tasks introduced in this dataset remains extremely relevant.

\begin{table}[H]
\begin{tabular}{l||c|c||c}
& Train & Test & Total\\
\hline
\# documents & 90.000 & 10.000& 100.000 \\
\hline
\# queries & 528.665 & 58.827& 587.492\\
\hline
\end{tabular}
\caption{Repartition of the questions into the train and test set.}
\label{train_test}
\end{table}
 
Then we randomly select 6 tasks for each review (the same task can be selected multiple times) and randomly select a natural language question that corresponds to this task. The questions are human-generated patterns that we have crowdsourced in order to produce a dataset as rich as possible. To this end, we have generated several patterns that correspond to the capabilities we wanted to express in a given question and we have crowdsourced rephrasing of these patterns.
\\

\begin{figure}
\includegraphics[scale=.5]{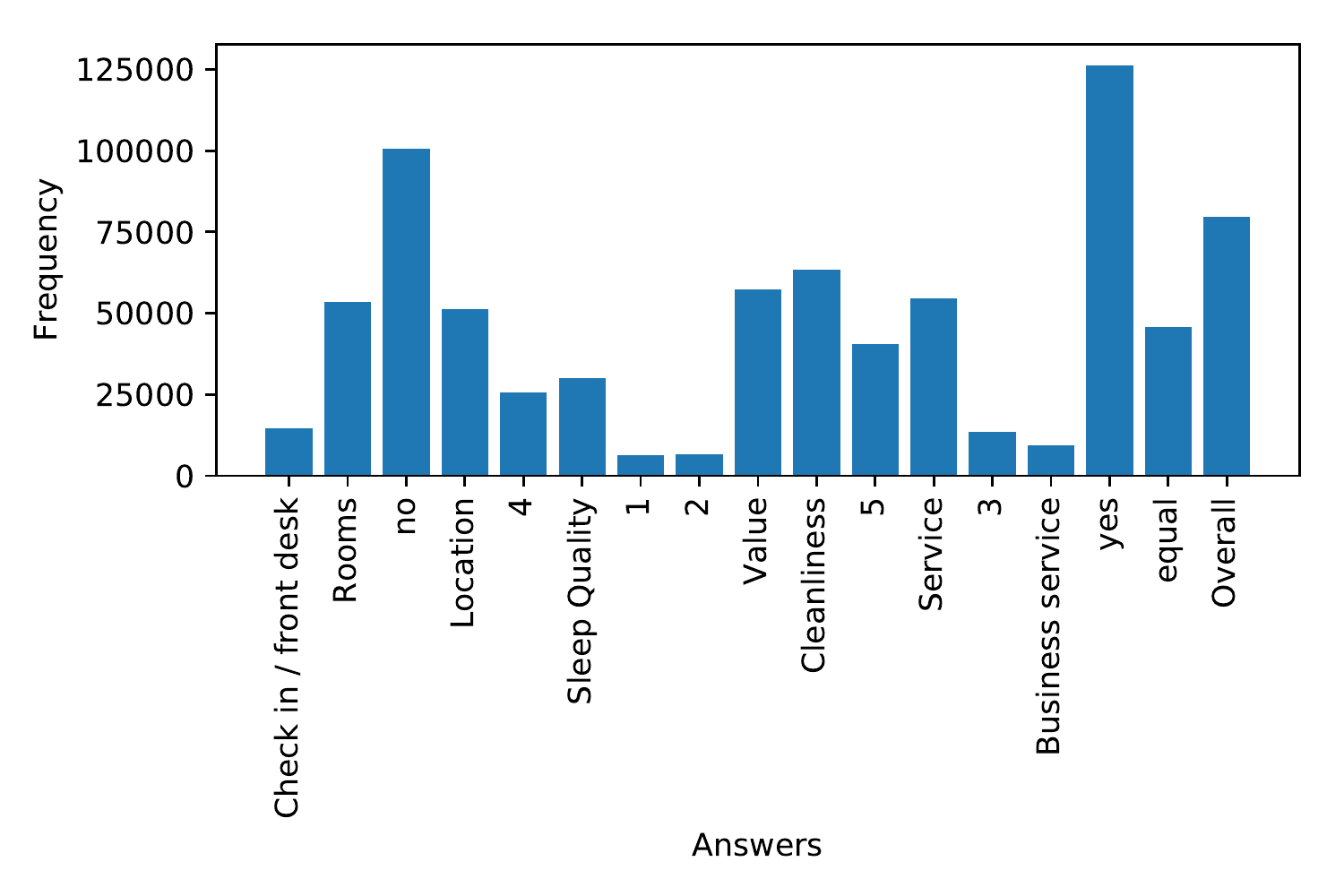}
\caption{Distribution of answers in the generated dataset.}
\label{answers}
\end{figure}

\begin{figure}[H]
\includegraphics[scale=.5]{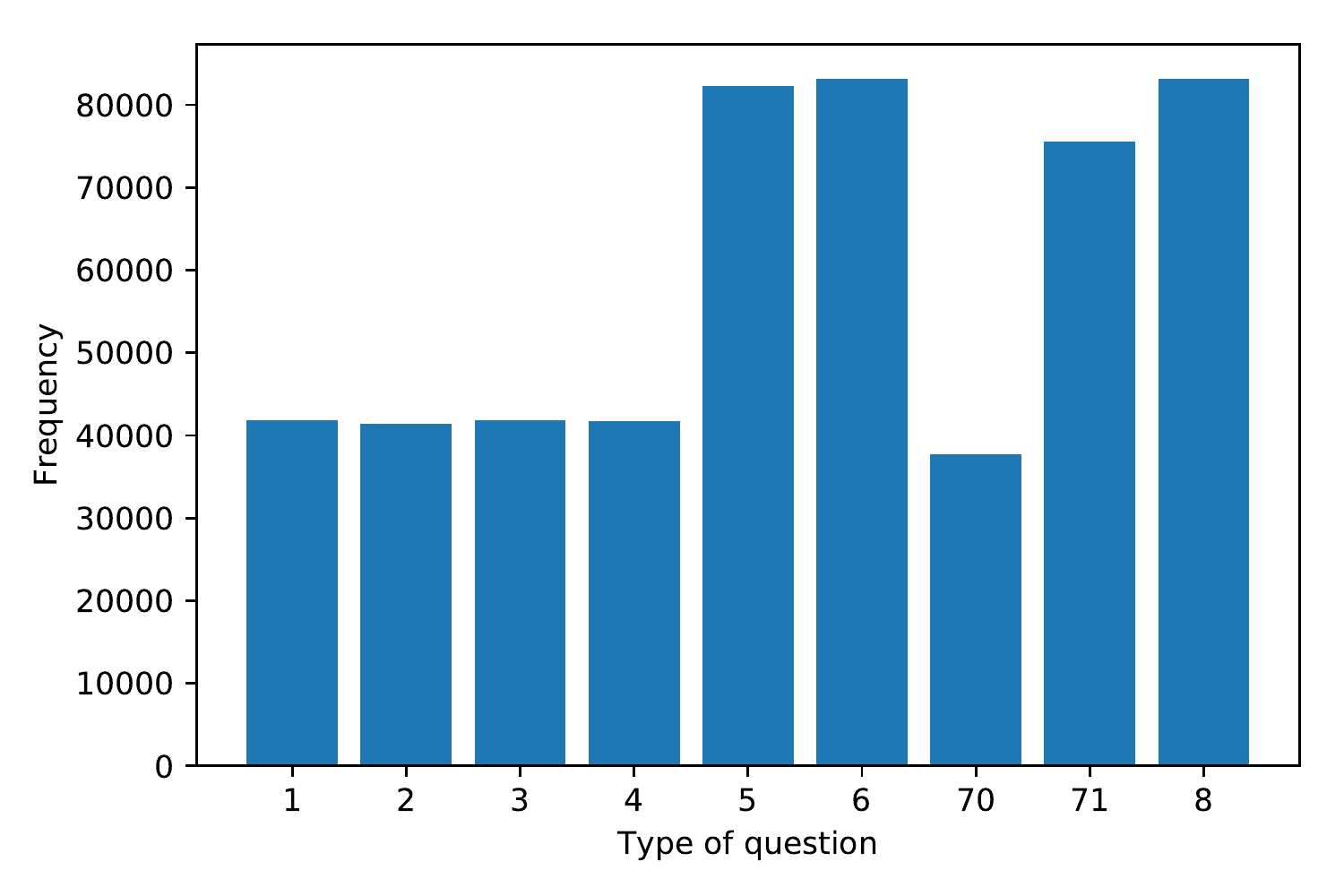}
\caption{Repartition of the tasks in the generated dataset.}
\label{types}
\end{figure}

The final dataset we propose is composed of more than 500.000 questions about 100.000 reviews. Table \ref{train_test} shows the repartition of the documents and queries into the train and test set.
Each review contains a maximum of 6 questions.
Sometimes less when it is not possible to generate all.
For example, if only two or three aspects are mentioned in a review, we will be able to generate only a little set of relational questions.
Figure \ref{answers} depicts the repartition of the answers in the generated dataset. A majority of the tasks we introduced, even if they possibly require a high level of understanding of the document and the question, are binary questions.
It means that in the generated dataset the answers \textit{yes} and \textit{no} tend to be more present than the others. 
To balance in a better way the distribution of the answers, we chose to affect a higher probability of sampling to the task 5, 6, 7.1, 8. Indeed, these tasks are not binary questions and required an aspect name as the answer. Figure \ref{types} represents the repartition of question types in our dataset. Finally, figure \ref{answers} shows the repartition of the answers in the dataset.

\subsection{Paraphrase augmentation using backtranslation}

In order to generate more paraphrases of the questions, we used a backtranslation method to enrich them. The idea is to use a translation model that will translate our human-generated questions into another language, and then translate them back to English.
This double translation will introduce rewordings of the questions that we will be able to integrate into this dataset.
This approach has been used in \cite{wei2018fast} to perform data augmentation on the training set. For this purpose, we have trained a fairseq \cite{gehring2016convenc} model to translate sentences from English to French and for French to English. In order to preserve the quality of the sentences we have so far, we only keep the most probable translation of each original sentence. Indeed a beam search is used during the translation to predict the most probable translations which mean that we each translation comes with an associated probability. By selecting only the first translations, we almost double the number of questions without degrading the quality of the questions proposed in the dataset.

\section{Experiments}
\label{sec:Experiments} 
\subsection{Models}

\begin{table*}
\begin{center}
\begin{tabular}{l|*{4}{c|}}

\hline
\multicolumn{1}{|c|}{\backslashbox{Task}{Model}}&\makebox[6.5em]{LogReg}&\makebox[6.5em]{LSTM}&\makebox[6.5em]{\memnn}&\makebox[6.5em]{Deep Proj reader}\\\hline\hline
\multicolumn{1}{|c|}{Overall} &  46.7 & 19.5 & 20.7 & \textbf{60.4} \\\hline\hline
\multicolumn{1}{|c|}{1} & 51.0 & 20.0 & 23.2 & \textbf{82.3}  \\\hline
\multicolumn{1}{|c|}{2} & 80.6 & 65.3 & 70.3 & \textbf{90.9} \\\hline
 \multicolumn{1}{|c|}{3}& 72.2 & 58.1 & 61.4 & \textbf{85.9} \\\hline
\multicolumn{1}{|c|}{4} & 58.4 & 28.1 & 28.0 & \textbf{91.3} \\\hline
\multicolumn{1}{|c|}{5} & 37.8 & 6.1 & 5.2 & \textbf{57.1} \\\hline
\multicolumn{1}{|c|}{6} & 16.0 & 8.3 & 10.1 & \textbf{39.1} \\\hline
\multicolumn{1}{|c|}{7} & 57.2 & 12.8 & 13.2 & \textbf{68.8} \\\hline
\multicolumn{1}{|c|}{8} & 36.8 & 18.0 & 17.8 & \textbf{41.3} \\\hline

\end{tabular}
\caption{Accuracy (\%) of 4 models on the ReviewQA dataset.}
\end{center}
\label{results}
\end{table*}

In this section, we present the performance of four different models on our dataset: a logistic regression and three neural models. The first one is a basic LSTM \cite{Hochreiter1997LongSM}, the second a \memnn \cite{Sukhbaatar2015EndToEndMN} and the third one is a model of our own design. This fourth model reuses the encoding layers of the R-net \cite{Wang2017GatedSN} and we modify the final layers with a projection layer that will be able to select the answer among the set of candidates instead of pointing the answerer directly into the source document.\\

\textbf{Logistic regression: }To produce the representation of the input, we concatenate the Bag-Of-Words representation of the document with the Bag-Of-Words representation of the question. It produces an array of size $2*V$ where $V$ is the vocabulary size. Then we use a logistic regression to select the most probable answer among the $N$ possibilities.\\

\textbf{LSTM: } We start with a concatenation of the sequence of indexes of the document with the sequence of indexes of the question. Them we feed an LSTM network with this vector and use the final state as the representation of the input. Finally, we apply a logistic regression over this representation to produce the final decision.\\

\textbf{End-to-end memory networks: }This architecture is based on two different memory cells (input and output) that contain a representation of the document. A controller, initialized with the encoding of the question, is used to calculate an attention between this controller and the representation of the document in the input memory. This attention is them used to re-weight the representation of the document in the output memory. This response from the output memory is them utilized to update the controller. After that, either a matrix is used to project this representation into the answer space either the controller is used to go through an over hop of memory. This architecture allows the model to sequentially look into the initial document seeking for important information regarding the current state of its controller. This model achieves very good performances on the 20 bAbI tasks dataset.\\

\textbf{Deep projective reader: }This is a model of our own design, largely inspired by the efficient R-net reader \cite{Wang2017GatedSN}. The overall architecture is composed of 4 stacked layers: an encoding layer, a question/document attention, a self-attention layer and a projection layer. The following paragraphs briefly describe the overall utility of each of these layers.

\begin{itemize}
\item \textbf{Encoding:} The sentence is tokenized by words. Each token is represented by the concatenation of its embedding vector and the final state of a bidirectional recurrent network over the characters of this word. Finally, another bidirectional RNN on the top of this representation produce the encoding of the document and the question.

\item \textbf{Question/document attention}: We apply a question/document attention layer that matches the representation of the question with each token of the document individually to output an attention that gives more weight to the important tokens of the document regarding the question.

\item \textbf{Self-attention layer:} The previous layer has built a question-aware representation of the document. One problem with such representation is that form the moment each token has only a good knowledge of its closest neighbors. To tackle this problem, \cite{Wang2017GatedSN} have proposed to use a self-attention layer that matches each individual token with all the other tokens of the document. Doing that, each token is now \textit{aware} of a larger context.

\item \textbf{Output layer:} A bidirectional RNN is applied on the top of the last layer and we use its final state as the representation of the input. We use a projection matrix to project this representation into the answer space and select the most probable one
\end{itemize}

\subsection{Training details}

We propose to train these models on the entire set of tasks and them to measure the overall performance and the accuracy of each individual task. In all the models, we use the Adam optimizer \cite{Kingma2014AdamAM} with a learning rate of 0.01 and the batch size is set to 64.
All the parameter are initialized from a Gaussian distribution with mean 0 and a standard deviation of 0.01.
The dimension of the word embeddings in the projective deep reading model and the LSTM model is 300 and we use Glove pre-trained vectors (\cite{Pennington2014GloveGV}). We use a \memnn with 5 memory hops and a linear start of 5 epochs. 
The reviews are split by sentence and each memory block corresponds to one sentence. 
Each sentence is represented by its bag-of-word representation augmented with temporal encoding as it is suggested in \cite{Sukhbaatar2015EndToEndMN}.

\subsection{Model performance}

Table \ref{results} displays the performance of the 4 baselines on the ReviewQA's test set. These results are the performance achieved by our own implementation of these 4 models. According to our results, the simple LSTM network and the \memnn perform very poorly on this dataset. Especially on the most advanced reasoning tasks. Indeed, the task 5 which corresponds to the prediction of the exact rating of an aspect seems to be very challenging for these model. Maybe the tokenization by sentence to create the memory blocks of the \memnn, which is appropriated in the case of the bAbI tasks, is not a good representation of the documents when it has to handle human generated comments. However, the logistic regression achieves reasonable performance on these tasks, and do not suffer from catastrophic performance on any tasks. Its worst result comes on task 6 and one of the reason is probably that this architecture is not designed to predict a list of answers. On the contrary, the deep projective reader achieves encouraging on this dataset. It outperforms all the other baselines, with very good scores on the first fourth tasks. The question/document and document/document attention layers proposed in \cite{Wang2017GatedSN} seem once again to produce rich encodings of the inputs which are relevant for our projection layer.

\section{Conclusion}
\label{sec:Conclusion}

In this paper, we formalize the sentiment analysis task through the framework of machine reading and release ReviewQA, a relational question-answering corpus. This dataset allows evaluating a set of relational reasoning skills through natural language questions. It is composed of a large panel of human-generated questions. Moreover, we propose to augment the dataset with backtranslated reformulations of these questions.
Finally, we evaluate 4 models on this dataset, including a projective model of our own design that seems to be a strong baseline for this dataset.
We expect that this large dataset will encourage the research community to develop reasoning models and evaluate their models on this set of tasks.

\section*{Acknowledgment}

We thank Vassilina Nikoulina and Stéphane Clinchant for the help regarding the backtranslation rewording of the questions.

\newpage
\bibliography{cap2018}

\newcommand{\etalchar}[1]{$^{#1}$}
\begin{thebibliography}{JHvdM{\etalchar{+}}17}

\bibitem[CFWB17]{Chen2017ReadingWT}
Danqi Chen, Adam Fisch, Jason Weston, and Antoine Bordes.
\newblock Reading wikipedia to answer open-domain questions.
\newblock In {\em ACL}, 2017.

\bibitem[GAGD16]{gehring2016convenc}
Jonas Gehring, Michael Auli, David Grangier, and Yann~N Dauphin.
\newblock {A Convolutional Encoder Model for Neural Machine Translation}.
\newblock {\em ArXiv e-prints}, November 2016.

\bibitem[HKG{\etalchar{+}}15]{Hermann2015TeachingMT}
Karl~Moritz Hermann, Tom{\'a}s Kocisk{\'y}, Edward Grefenstette, Lasse
  Espeholt, Will Kay, Mustafa Suleyman, and Phil Blunsom.
\newblock Teaching machines to read and comprehend.
\newblock In {\em NIPS}, 2015.

\bibitem[HLL{\etalchar{+}}17]{DBLP:journals/corr/abs-1711-05073}
Wei He, Kai Liu, Yajuan Lyu, Shiqi Zhao, Xinyan Xiao, Yuan Liu, Yizhong Wang,
  Hua Wu, Qiaoqiao She, Xuan Liu, Tian Wu, and Haifeng Wang.
\newblock Dureader: a chinese machine reading comprehension dataset from
  real-world applications.
\newblock {\em CoRR}, abs/1711.05073, 2017.

\bibitem[HPQ17]{DBLP:journals/corr/HuPQ17}
Minghao Hu, Yuxing Peng, and Xipeng Qiu.
\newblock Mnemonic reader for machine comprehension.
\newblock {\em CoRR}, abs/1705.02798, 2017.

\bibitem[HS97]{Hochreiter1997LongSM}
Sepp Hochreiter and J{\"u}rgen Schmidhuber.
\newblock Long short-term memory.
\newblock {\em Neural computation}, 9 8:1735--80, 1997.

\bibitem[JHvdM{\etalchar{+}}17]{Johnson2017CLEVRAD}
Justin Johnson, Bharath Hariharan, Laurens van~der Maaten, Li~Fei-Fei,
  C.~Lawrence Zitnick, and Ross~B. Girshick.
\newblock Clevr: A diagnostic dataset for compositional language and elementary
  visual reasoning.
\newblock {\em 2017 IEEE Conference on Computer Vision and Pattern Recognition
  (CVPR)}, pages 1988--1997, 2017.

\bibitem[JL17]{Jia2017AdversarialEF}
Robin Jia and Percy Liang.
\newblock Adversarial examples for evaluating reading comprehension systems.
\newblock In {\em EMNLP}, 2017.

\bibitem[KB14]{Kingma2014AdamAM}
Diederik~P. Kingma and Jimmy Ba.
\newblock Adam: A method for stochastic optimization.
\newblock {\em CoRR}, abs/1412.6980, 2014.

\bibitem[LP17]{Liu2017GatedEM}
Fei Liu and Julien Perez.
\newblock Gated end-to-end memory networks.
\newblock In {\em EACL}, 2017.

\bibitem[NRS{\etalchar{+}}16]{Nguyen2016MSMA}
Tri Nguyen, Mir Rosenberg, Xia Song, Jianfeng Gao, Saurabh Tiwary, Rangan
  Majumder, and Li~Deng.
\newblock Ms marco: A human generated machine reading comprehension dataset.
\newblock {\em CoRR}, abs/1611.09268, 2016.

\bibitem[PSM14]{Pennington2014GloveGV}
Jeffrey Pennington, Richard Socher, and Christopher~D. Manning.
\newblock Glove: Global vectors for word representation.
\newblock In {\em EMNLP}, 2014.

\bibitem[RZLL16]{Rajpurkar2016SQuAD10}
Pranav Rajpurkar, Jian Zhang, Konstantin Lopyrev, and Percy Liang.
\newblock Squad: 100, 000+ questions for machine comprehension of text.
\newblock In {\em EMNLP}, 2016.

\bibitem[SSWF15]{Sukhbaatar2015EndToEndMN}
Sainbayar Sukhbaatar, Arthur Szlam, Jason Weston, and Rob Fergus.
\newblock End-to-end memory networks.
\newblock In {\em NIPS}, 2015.

\bibitem[TQL16]{Tang2016AspectLS}
Duyu Tang, Bing Qin, and Ting Liu.
\newblock Aspect level sentiment classification with deep memory network.
\newblock In {\em EMNLP}, 2016.

\bibitem[TTH17]{Tay2017DyadicMN}
Yi~Tay, Luu~Anh Tuan, and Siu~Cheung Hui.
\newblock Dyadic memory networks for aspect-based sentiment analysis.
\newblock In {\em CIKM}, 2017.

\bibitem[TWY{\etalchar{+}}17]{Trischler2017NewsQAAM}
Adam Trischler, Tong Wang, Xingdi Yuan, Justin Harris, Alessandro Sordoni,
  Philip Bachman, and Kaheer Suleman.
\newblock Newsqa: A machine comprehension dataset.
\newblock In {\em Rep4NLP@ACL}, 2017.

\bibitem[WBCM15]{Weston2015TowardsAQ}
Jason Weston, Antoine Bordes, Sumit Chopra, and Tomas Mikolov.
\newblock Towards ai-complete question answering: A set of prerequisite toy
  tasks.
\newblock {\em CoRR}, abs/1502.05698, 2015.

\bibitem[WLZ10]{Wang2010LatentAR}
Hongning Wang, Yue Lu, and ChengXiang Zhai.
\newblock Latent aspect rating analysis on review text data: a rating
  regression approach.
\newblock In {\em KDD}, 2010.

\bibitem[WLZ11]{Wang2011LatentAR}
Hongning Wang, Yue Lu, and ChengXiang Zhai.
\newblock Latent aspect rating analysis without aspect keyword supervision.
\newblock In {\em KDD}, 2011.

\bibitem[WSR17]{Welbl2017ConstructingDF}
Johannes Welbl, Pontus Stenetorp, and Sebastian Riedel.
\newblock Constructing datasets for multi-hop reading comprehension across
  documents.
\newblock {\em CoRR}, abs/1710.06481, 2017.

\bibitem[WYW{\etalchar{+}}17]{Wang2017GatedSN}
Wenhui Wang, Nan Yang, Furu Wei, Baobao Chang, and Ming Zhou.
\newblock Gated self-matching networks for reading comprehension and question
  answering.
\newblock In {\em ACL}, 2017.

\bibitem[YDL{\etalchar{+}}18]{wei2018fast}
Adams~Wei Yu, David Dohan, Quoc Le, Thang Luong, Rui Zhao, and Kai Chen.
\newblock Fast and accurate reading comprehension by combining self-attention
  and convolution.
\newblock In {\em International Conference on Learning Representations}, 2018.

\end{thebibliography}

\end{document}